\documentclass[conference]{IEEEtran}
\IEEEoverridecommandlockouts
\usepackage{amsmath,amssymb,amsfonts}
\usepackage{algorithmic}
\usepackage{graphicx}
\usepackage{textcomp}
\usepackage{xcolor}

\usepackage{pifont}
\newcommand{\cmark}{\ding{51}} 
\newcommand{\xmark}{\ding{55}} 
\renewcommand{\arraystretch}{1.05}
\usepackage{booktabs}
\usepackage{makecell}  
\usepackage[style=ieee]{biblatex}
\usepackage{multirow}
\usepackage{siunitx}
\usepackage{indentfirst}
\usepackage{comment}
\usepackage{enumitem}
\usepackage{float}
\usepackage{graphicx}
\usepackage{wrapfig}
\usepackage{adjustbox}
\usepackage{svg}
\usepackage{hyperref}
\usepackage{titlesec}

\renewcommand{\comment}[1]{{\color{darkgray} // #1}}

\title{FPG-NAS: FLOPs-Aware Gated Differentiable Neural Architecture Search for Efficient 6DoF Pose Estimation}

\author{
  \IEEEauthorblockN{%
    Nassim {Ali Ousalah}\IEEEauthorrefmark{1}\quad
    Peyman Rostami\IEEEauthorrefmark{1}\quad
    Anis Kacem\IEEEauthorrefmark{1}\quad
    Enjie Ghorbel\IEEEauthorrefmark{1}\IEEEauthorrefmark{2}\\[0.3em]
    Emmanuel Koumandakis\IEEEauthorrefmark{3}
    Djamila Aouada\IEEEauthorrefmark{1}
  }
  \vspace{0.5em}
   \IEEEauthorblockA{\IEEEauthorrefmark{1}SnT, University of Luxembourg, Luxembourg\\\IEEEauthorrefmark{2}Cristal Lab, ENSI, Manouba University, Tunisia\\\IEEEauthorrefmark{3}Infinite Orbits, Toulouse, France}
   \IEEEauthorblockA{\texttt{\{nassim.aliousalah, peyman.rostami, anis.kacem, djamila.aouada\}@uni.lu} \\ \texttt{enjie.ghorbel@isamm.uma.tn} \\ \texttt{manos@infiniteorbits.io}}
   \vspace{-5mm}
   }

\def\BibTeX{{\rm B\kern-.05em{\sc i\kern-.025em b}\kern-.08em
    T\kern-.1667em\lower.7ex\hbox{E}\kern-.125emX}}
\begin{document}



\maketitle
\vspace{-1em}

\begin{abstract}

    We introduce FPG-NAS, a FLOPs-aware Gated Differentiable Neural Architecture Search framework for efficient 6DoF object pose estimation. Estimating 3D rotation and translation from a single image has been widely investigated yet remains computationally demanding, limiting applicability in resource-constrained scenarios. FPG-NAS addresses this by proposing a specialized differentiable NAS approach for 6DoF pose estimation, featuring a task-specific search space and a differentiable gating mechanism that enables discrete multi-candidate operator selection, thus improving architectural diversity. Additionally, a FLOPs regularization term ensures a balanced trade-off between accuracy and efficiency. The framework explores a vast search space of approximately 10\textsuperscript{92} possible architectures. Experiments on the LINEMOD and SPEED+ datasets demonstrate that FPG-NAS-derived models outperform previous methods under strict FLOPs constraints. To the best of our knowledge, FPG-NAS is the first differentiable NAS framework specifically designed for 6DoF object pose estimation.
\end{abstract}


\section{Introduction}
\label{sec:intro}

The problem of Six Degrees of Freedom (6DoF) pose estimation is a fundamental one in computer vision, with a wide range of applications in robotics, augmented reality (AR), space exploration, and autonomous navigation systems~\cite{ornek2024foundpose, Nguyen_2024_CVPR, xu20246d, PAULY2023339}. The task involves estimating the 3D orientation (rotation) and position (translation) of an object with respect to the camera coordinate system.


Despite significant progress, 6DoF pose estimation methods continue to face challenges when deployed on resource-constrained devices. Many existing approaches rely on standard or off-the-shelf backbone networks, which often incur high computational costs (FLOPs) and substantial latency~\cite{labbe2020cosypose, Su_2022_CVPR, he2020pvn3d, park2024robust}. This highlights the need for more computationally efficient architectures tailored for 6DoF pose estimation.

\begin{figure}[t]
    \includegraphics[width=0.475\textwidth]{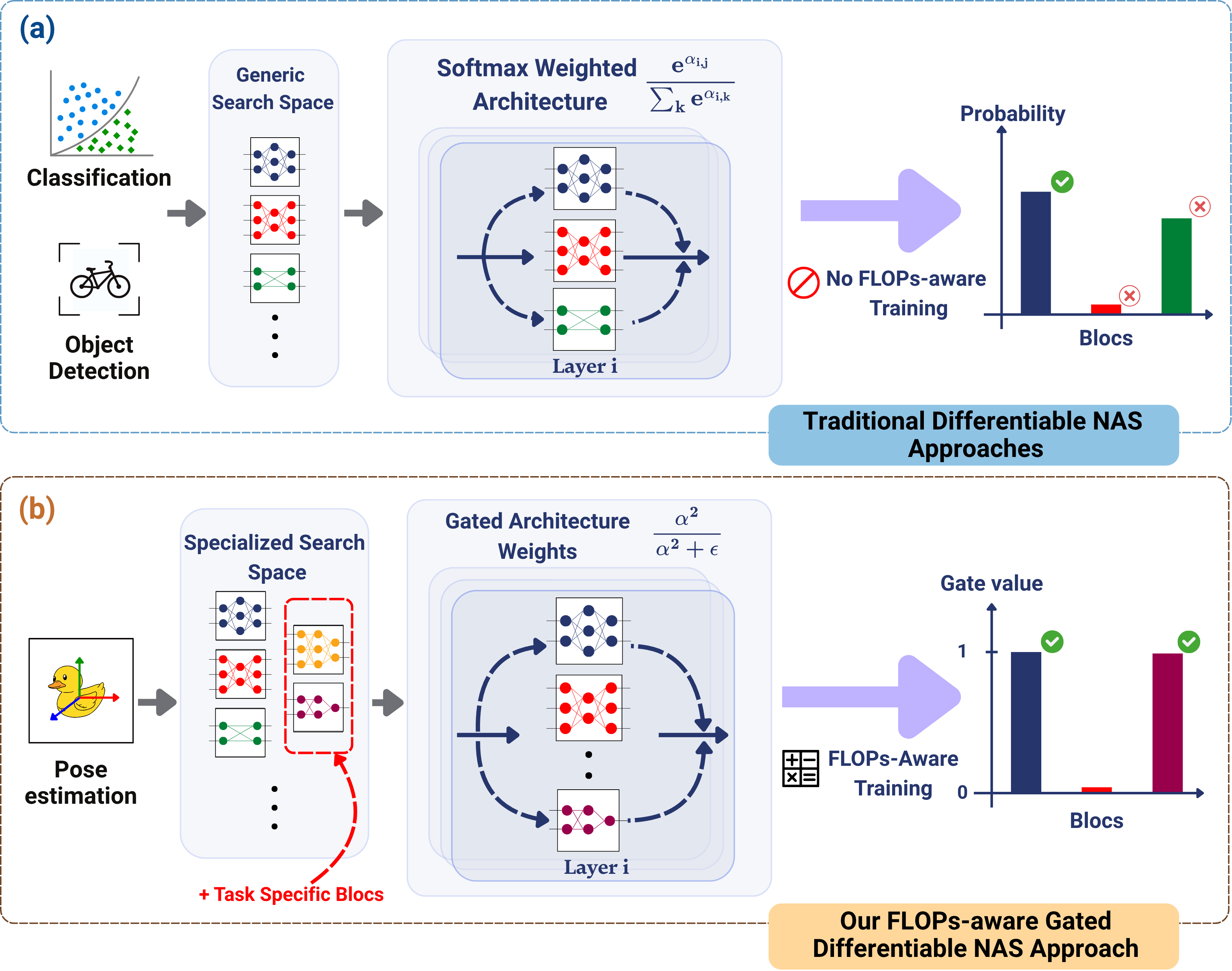}
    \caption{Traditional DNAS vs \textbf{FPG-NAS}. (a) Traditional DNAS uses \texttt{softmax} weighting potentially overlooking valuable block combinations. (b) \textbf{FPG-NAS} incorporates task-specific blocks and a gated mechanism that enables the selection of multiple synergistic blocks.}
    \label{fig:motivation}
    \vspace{-4mm}
\end{figure}

Several studies have attempted to address efficiency concerns by compressing deep neural networks or designing lightweight architectures~\cite{Rostami_2025_WACV, Sinha_2024_WACV, Guo_2021_ICCV, chee2023quip, ousalah2025uncertainty, sinha2024multi}. Among these efforts, Neural Architecture Search (NAS), and more specifically Differentiable NAS (DNAS), has emerged as a promising direction. DNAS~\cite{liu2018darts,cai2018proxylessnas, xie2018snas, wu2019fbnet, zhu2023improving, che2022differentiable} provides a powerful framework for automating network design by formulating architecture search as a continuous optimization problem, allowing the use of gradient-based optimization techniques. 



Nevertheless, directly extending standard DNAS approaches to the task of 6DoF pose estimation is challenging for two main reasons. First, although DNAS has achieved notable success in tasks such as image classification and object detection, its application to 6DoF pose estimation, particularly in the 2-stage setting of keypoint detection followed by PnP~\cite{chen2022epro, item_34721c4a90c2421c9df615a96127a4a2}, remains largely unexplored. This is important, as 6DoF pose estimation typically requires a larger overall search space, the integration of task-specific operations, and design biases that are often absent in conventional DNAS frameworks~\cite{liu2018darts, wu2019fbnet, xie2018snas, guo2020single}.

Second, DNAS methods usually discretize their continuous relaxation by fixing each network's layer to the operator with the highest \texttt{softmax} probability via an \texttt{argmax} step~\cite{liu2018darts}. While this strategy is suitable in other domains, it can be particularly limiting for pose estimation, where combining multiple operations across different stages of the network may be crucial for capturing the complex spatial relationships inherent in 6DoF tasks. This standard “winner-takes-all” approach may discard useful block combinations, as illustrated in Figure~\ref{fig:motivation}(a).

To overcome this, we enable the network to combine multiple block operations per layer via differentiable gating, allowing smooth optimization and gradual sparsification without abrupt \texttt{argmax} decisions. Originally designed for channel pruning~\cite{Guo_2021_ICCV}, we repurpose gating to guide candidate selection during architecture search. As shown in Figure~\ref{fig:motivation}(b), this preserves multiple promising candidates, expands the search space, and enhances exploration through discretized multi-candidate selection.


Motivated by these insights, we propose \textbf{FPG-NAS}, a \textit{\textbf{F}LO\textbf{P}s-aware \textbf{G}ated Differentiable \textbf{NAS} framework} for efficient 6DoF pose estimation. Our main contributions can be summarized as follows: 

\begin{itemize}
    \item A task-specific search space designed for the 6DoF pose estimation problem, comprising different FLOPs-efficient operations and supporting approximately $10^{92}$ possible architectures.
    \item A solution based on differentiable gating mechanisms that enables flexible, per-layer selection of multiple block operations.
    
    \item Extensive empirical validation on the LINEMOD~\cite{10.1007/978-3-642-37331-2_42} and SPEED+~\cite{park2022speed+} datasets, demonstrating superior performance and significantly improved efficiency compared to state-of-the-art methods.
\end{itemize}

\section{Related Work}
\label{sec:relatedwork}

\subsection{Differentiable Neural Architecture Search}
Differentiable Neural Architecture Search (DNAS) has shown success in standard vision tasks such as classification and detection~\cite{liu2018darts, xie2018snas, cai2018proxylessnas, wu2019fbnet}, but remains underexplored for more complex problems such as 6DoF pose estimation, which may require task-adapted search spaces. Additionally, not all DNAS methods include computational constraints during optimization, yielding architectures that are accurate but inefficient. Another key limitation lies in the \texttt{softmax}-based operation selection~\cite{liu2018darts}, which assumes a sharply peaked distribution. When multiple operations have similar scores, \texttt{argmax} can discard viable alternatives, introducing a discretization gap. While recent methods like SD-DARTS~\cite{zhu2023improving}, SA-DARTS~\cite{zhou2025regularizing}, and sparse DARTS~\cite{zhang2021robustifying} attempt to reduce this gap through regularization or distillation, they still retain the same limited search space and do not integrate FLOPs or latency into the optimization process.

\subsection{Efficient 6DoF Pose Estimation}
Recent research on efficient 6DoF object pose estimation has focused on reducing computational cost through lightweight architectures~\cite{Kehl_2017_ICCV, bukschat2020efficientpose}, quantization~\cite{javed2023modular}, and knowledge distillation~\cite{guan2022hrpose, Guo_2023_CVPR, ousalah2025uncertainty}. Approaches include using computationally efficient backbones such as SSD~\cite{Kehl_2017_ICCV} and EfficientDet~\cite{bukschat2020efficientpose}, compressing networks via Modular Quantization-Aware Training~\cite{javed2023modular}, and training compact student models to replicate the output of larger teacher models while maintaining accuracy. Despite these advancements, all existing efficient architectures have been designed manually. Differentiable Neural Architecture Search (DNAS) remains largely unexplored in this domain, presenting an opportunity to develop more principled and task-specific networks for 6DoF pose estimation.

\section{Proposed Approach}
\label{sec:metodology}

In this section, we provide details on \textbf{FPG-NAS}, including the micro and macro architectures of our search space, the FLOPs-aware loss, and the proposed search algorithm.

\subsection{Search Space} \label{subsec:search_space}
Following prior works~\cite{wu2019fbnet, liu2018darts}, our search space allows a layer-wise search with a fixed macro architecture.


\noindent \textbf{Macro Architecture:} As outlined in Section~\ref{sec:intro}, we adopt a keypoint-based pose estimation approach. Our macro architecture consists of 17 backbone layers, a multi-scale fusion layer integrating feature maps from different stages, and a heatmap head for regressing the keypoints. To adapt the search for keypoint detection, where spatial precision and multi-scale reasoning are crucial, we include the fusion layer in the search space. Unlike prior methods with fixed fusion strategies (e.g., DeConvolution, concatenation and FPN~\cite{xiang2017posecnn, Wang_2019_CVPR, wang2021gdr, Peng_2019_CVPR}), we enable the fusion mechanism itself to be searched. Other components (stem, depthwise separable head, input-output resolutions) are fixed to reduce search complexity. In total, 16 backbone layers and the fusion layer are searchable. Output channels increase with network depth to help manage FLOPs, which scale quadratically with resolution. The full layer configuration is provided in Table~\ref{tab:macro_arch}.

\begin{table}[ht]
    \centering
    \small 
    \setlength{\tabcolsep}{3pt}
    \caption{\textbf{Macro architecture.} Stem, 16 \textit{SB} layers, multi-scale fusion, and head predicting $\text{K}$ keypoints. \textbf{s}: stride, \textbf{DW-Conv}: depthwise conv, \textit{\textbf{SB}}: searchable block.}
    \label{tab:macro_arch}
    \begin{tabular}{p{1.5cm}cccccc}
        \hline
        \textbf{Stage} & \textbf{Input Shape} & \textbf{Block type} & $\textbf{\#}\textbf{\text{C}}_{\textbf{\text{out}}}$ & $\textbf{\#}\textbf{L}$ & \textbf{s} \\
        \hline
        \textcolor{white}{-}\comment{Stem} & $W \times H \times 3$ & $3\times3$ Conv & 32 & 1 & 2 \\
         & $W/2 \times H/2 \times 32$ & \textit{SB} & 16 & 1 & 1 \\
        \textcolor{white}{-}\comment{S1} & $W/2 \times H/2 \times 16$ & \textit{SB} & 24 & 1 & 2 \\
        \textcolor{white}{-}\comment{S2} & $W/4 \times H/4 \times 24$ & \textit{SB} & 40 & 2 & 2 \\
         & $W/8 \times H/8 \times 40$ & \textit{SB} & 80 & 2 & 2 \\
        \textcolor{white}{-}\comment{S4} & $W/16 \times H/16 \times 80$ & \textit{SB} & 112 & 3 & 1 \\
         & $W/16 \times H/16 \times 112$ & \textit{SB} & 192 & 3 & 2 \\
        \textcolor{white}{-}\comment{S6} & $W/32 \times H/32 \times 192$ & \textit{SB} & 320 & 4 & 1 \\
        \hline
        \textcolor{white}{-}\comment{Fusion} & S\{1,2,4,6\} & \textit{SB} & 64 & 1 & - \\
        \hline
        \textcolor{white}{-}\comment{Head} & $W/4 \times H/4 \times 64$ & DW-Conv & $\text{K}$ & 1 & - \\
        \hline
    \end{tabular}
\end{table}

\begin{figure}[ht]
    \centering
    \includegraphics[width=0.4\textwidth]{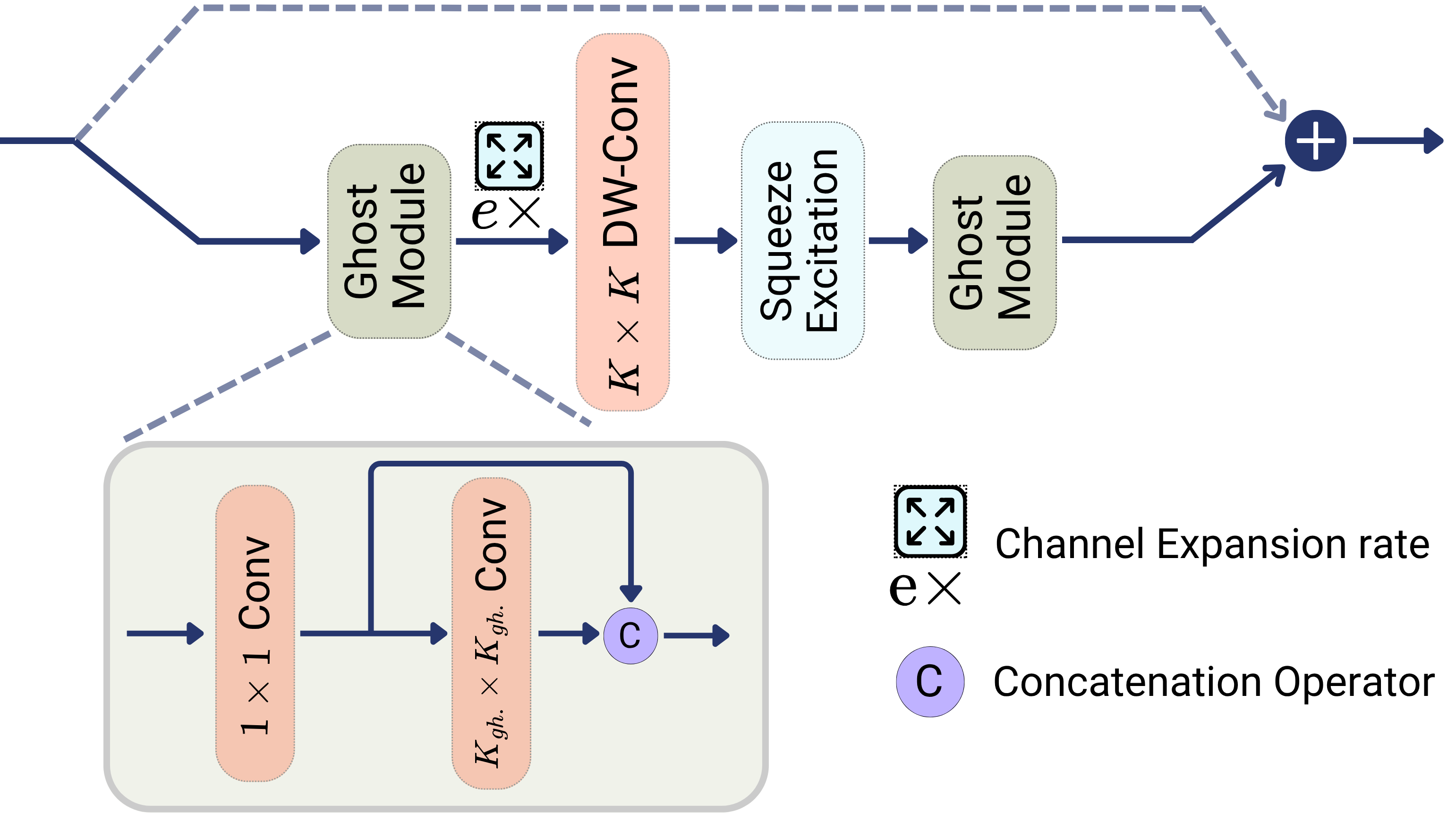}

    \caption{\textbf{Single backbone block.} Each candidate selects a kernel size~$K$, an expansion rate~$e$, and a ghost module kernel size~$K_{gh}$.}

    \label{fig:micro_arch}
    \vspace{-1em}
\end{figure}

\noindent\textbf{Micro Architecture:} Our micro architecture, shown in Figure~\ref{fig:micro_arch}, is inspired by EfficientNet~\cite{tan2019efficientnet} and GhostNet~\cite{han2020ghostnet}.  Our bottleneck block begins with a Ghost Module (ghost ratio of $2$) to efficiently expand the input channels, followed by a depthwise convolution, batch normalization, and SiLU activation, then a Squeeze-and-Excitation layer and finally a second Ghost Module. A residual connection is included when the input and output dimensions match. We denote the kernel size of the main depthwise convolution as $K$, the expansion rate as $e$, and the kernel size within the Ghost module as $K_{\text{ghost}}$. These parameters trade-off between local/global spatial modeling, representational capacity, and computational efficiency across layers. Specifically, $K \in \{3, 5, 7\}$, $e \in \{1, 3, 6\}$, and $K_{\text{ghost}} \in \{3, 5\}$. With an optional skip connection, each backbone layer selects from $3 \times 3 \times 2 + 1 = 19$ candidate blocks.


For the fusion layer, we introduce a dedicated search space with four FLOPs-efficient candidates that operate on identical multi-scale feature maps and output at a unified resolution, but each offer a distinct cross-scale interaction strategy:  (1)~\textit{SimpleFusion} uses $1\times1$ convolutions for lightweight feature merging. (2)~\textit{DilatedFusion} expands spatial context by increasing the receptive field using dilated $3\times3$ convolutions. (3)~\textit{SEFusion} applies channel-wise feature recalibration using $1\times1$ projections and Squeeze-and-Excitation. (4)~\textit{AttentionFusion} projects and merges features with $1\times1$ convolutions, then enhances salient spatial regions using a spatial attention mask derived from pooled spatial statistics. This diversity enables the search to explore trade-offs between efficiency and expressivity.

\subsection{The Search Algorithm}
Differentiable Neural Architecture Search (DNAS) frames architecture search as a bilevel optimization problem, enabling efficient gradient-based optimization via continuous relaxation of the discrete search space. The goal of DNAS is to jointly optimize network weights $w$ and architecture parameters $\alpha$ by solving:

\vspace{-5mm}
\begin{equation} \label{eq:dnas}
\begin{array}{c}
    \displaystyle \min_{\alpha} \ \  \mathcal{L}_{\text{val}}(w^*(\alpha), \alpha) \\ [1em]
    \text{s.t.} \quad  w^*(\alpha) = \displaystyle \arg\min_{w} \mathcal{L}_{\text{train}}(w, \alpha)
\end{array}
\end{equation}
\vspace{-3mm}

\noindent where $\mathcal{L}_{\text{train}}$ and $\mathcal{L}_{\text{val}}$ are the training loss and the validation loss, respectively. However, traditional approaches~\cite{liu2018darts, cai2018proxylessnas, wu2019fbnet, xie2018snas, zhu2023improving} often neglect combinations of operators that could synergistically cooperate to improve performance. To address this, we employ Differentiable Polarized Gates~\cite{Guo_2021_ICCV}, allowing smooth optimization while effectively steering the architecture parameters towards discrete decisions. Each candidate block in the architecture is controlled by a differentiable gate defined as,

\vspace{-2mm}
\begin{equation}
g(\alpha) = \frac{\alpha^2}{\alpha^2 + \epsilon} =
\begin{cases}
0, & \alpha = 0 \\[0.5em]
\approx 1, & \alpha \neq 0
\end{cases}
\end{equation}
\vspace{-2mm}

\noindent where $\epsilon$ is a small constant that controls the sharpness of the gate. This formulation allows gates to smoothly approximate discrete binary choices ($0$ or $1$), and allows multiple candidate blocks to remain simultaneously active within each layer, substantially expanding the effective search space. Specifically, each of the $16$ layers can select any non-empty subset from the $19$ candidates, yielding $2^{19}-1$ choices per layer. Incorporating the $4$ fusion layer options, the search space reaches \( 4 \times (2^{19} - 1)^{16} \approx 10^{92} \) possible architectures.

For fusion layers only, we train with a standard Gumbel-Softmax~\cite{wu2019fbnet, jang2016categorical}. With four candidates and a strict FLOPs budget, the distribution is naturally sparse and discrete. All architecture parameters $\alpha$ are initialized to $1$, with $\epsilon$ set to $0.1$ and gradually decayed during training to improve gradient stability. The output of layer $l$ is then computed as:

\vspace{-2mm}
\begin{equation}
    x_{l+1} = \sum^{N}_{i=1} g(\alpha_i) \cdot b_{l,i}(x_l)
\end{equation}
\noindent where $b_{l,i}(x_l)$ is the output of block $i$ in layer $l$, and $N = 19$ as defined in Section~\ref{subsec:search_space}.

\begin{figure}[t]
\centering
    \includegraphics[width=0.475\textwidth]{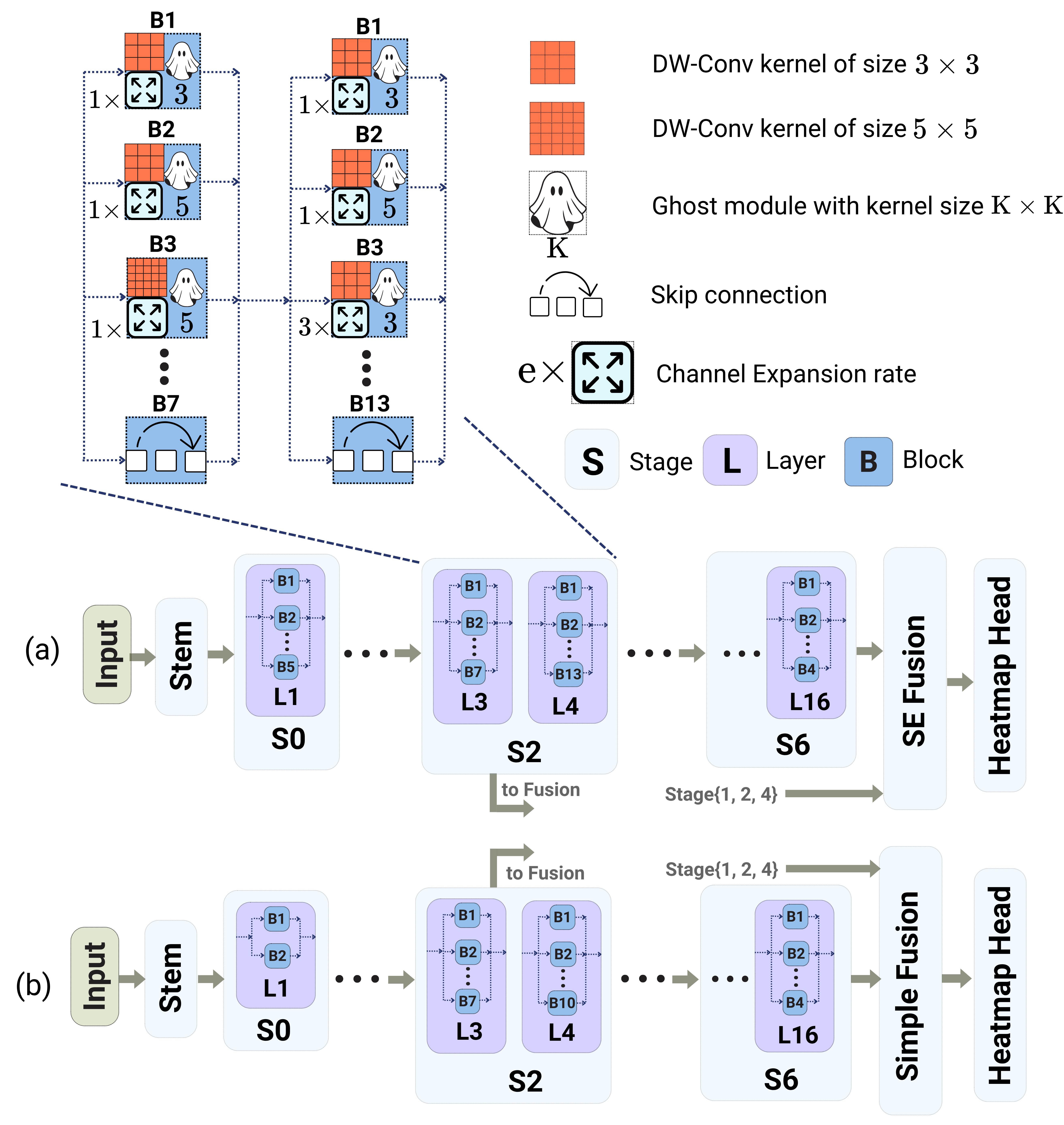}
    \caption{\textbf{Architectures discovered by FPG-NAS} under different FLOPs constraints: (a) 20 GFLOPs, (b) 10 GFLOPs. The figure illustrates the multi-block selections per layer and the fusion strategy chosen by the search.}
    \label{fig:discovered_arch}
\end{figure}

\noindent \textbf{FLOPs regularization term.} The supernet is trained by minimizing the following overall loss:
\vspace{-2mm}
\begin{equation} \label{eq:overall_loss}
    \mathcal{L}_{\text{train}} = \mathcal{L}_{\text{task}} + \lambda \cdot \text{ReLU} (\text{FLOPs}_{\text{total}} - \text{FLOPs}_{\text{budget}})
\end{equation}

\noindent where $\mathcal{L}_{\text{task}}$ is the task-specific loss, $\text{FLOPs}_{\text{budget}}$ is a user-defined target computational cost, $\text{ReLU}$ is the rectified linear unit function, and $\lambda$ controls the strength of the FLOPs regularization term. The regularization term penalizes architectures that exceed the target FLOPs budget. The total computational cost $\text{FLOPs}_{\text{total}}$ is computed by weighting the FLOPs of each candidate operation by its corresponding gate value (backbone) or architecture parameter (fusion layer):
\begin{equation}
\begin{aligned}
\text{FLOPs}_{\text{total}}
  &= \sum_{l,i} g(\alpha_{l,i})\,\text{FLOPs}_{l,i}
     + \sum_k \alpha_k\,\text{FLOPs}_k \\[2pt]
  &\quad + \text{FLOPs}_{\text{stem+head}}
\end{aligned}
\end{equation}

\noindent where $\text{FLOPs}_{l,i}$ and $\text{FLOPs}_k$ are the costs of block $i$ in layer $l$ and fusion layer $k$, respectively. In all experiments, FLOPs are reported in billions and $\lambda$ increases linearly from $0.1$ to $1$ during training. Figure~\ref{fig:discovered_arch} illustrates an overview of the architectures discovered by \textbf{FPG-NAS}. The search consistently selects multiple blocks per layer and optimizes the fusion strategy jointly.



\section{Experiments}
\label{sec:experiments}

\subsection{Experimental Setup}

We evaluate \textbf{FPG-NAS} on two 6DoF pose estimation datasets, LINEMOD~\cite{10.1007/978-3-642-37331-2_42} and SPEED+~\cite{park2022speed+}. LINEMOD comprises 16,000 images across 13 object categories. SPEED+ contains 60,000 synthetic images of the Tango spacecraft~\cite{amico2014tango} for training, with two other test subsets \texttt{Lightbox} and \texttt{Sunlamp}. Though not a standard 6DoF benchmark, SPEED+ is crucial for space applications requiring efficient models deployable on resource-limited hardware.

\subsection{Datasets \& Preprocessing}

For LINEMOD~\cite{10.1007/978-3-642-37331-2_42}, we follow standard practice~\cite{Su_2022_CVPR, Peng_2019_CVPR, xu20246d} by cropping training images around the target object and applying the same augmentation pipeline as in~\cite{Guo_2023_CVPR}.  For SPEED+~\cite{park2022speed+}, architecture search is conducted on a proxy set (one-third of the training data), and the selected models are trained on the full dataset using the augmentation strategy described in~\cite{park2024robust}.

\subsection{Training \& Hyper-parameters}

For LINEMOD, the task loss in Equation~(\ref{eq:overall_loss}) is the Kullback-Leibler (KL) divergence. For SPEED+, we use the mean squared error (MSE). All models are optimized with AdamW. The architecture weights are trained with a cosine-decay learning rate schedule, decaying from $10^{-2}$ to $10^{-4}$, while a fixed learning rate of $10^{-3}$ is used for the remaining weights.

Architecture search is conducted for $20$ epochs on LINEMOD and $30$ epochs on SPEED+. The discovered architectures are retrained on the full datasets for the same number of epochs. The search time on an NVIDIA A100 GPU is approximately $0.21$ GPU-days for LINEMOD (per class) and $1.8$ GPU-days for SPEED+. During training, we employ automatic mixed precision (AMP)~\cite{micikevicius2017mixed} with dynamic loss scaling for better stability. We also apply gradient clipping with a maximum norm of $1.0$ to prevent gradient explosion.

\subsection{Baselines} 
We compare architectures discovered by \textbf{FPG-NAS} with several existing manually designed methods that report the FLOPs of their models.
To the best of our knowledge, we are the first to apply DNAS to the task of  6DoF object pose estimation. Additionally, for a thorough comparison, we implement the commonly used \texttt{softmax}-\texttt{argmax} weighting strategy of DARTS~\cite{liu2018darts} and use it as a comparison baseline.
We also replicate the DNAL~\cite{GUO2022108448} method, where architecture parameters are controlled by a scaled \texttt{sigmoid} function that behaves similarly to a gating mechanism. We conduct architecture search using \textbf{FPG-NAS} on LINEMOD and SPEED+ datasets under FLOPs budgets of 10 and 5 GFLOPs.  We denote the resulting models as \textbf{Ours$_{\text{A}}$} and \textbf{Ours$_{\text{B}}$}, corresponding to the architectures found with 10 and 5 GFLOPs budgets, respectively. 

\begingroup
  \setlength{\tabcolsep}{4pt}        
  \renewcommand{\arraystretch}{1.1} 
  \setlength{\aboverulesep}{0pt}     
  \setlength{\belowrulesep}{0pt}     
  
  \begin{table*}[ht]
    \centering
    \small
    \caption{\textbf{Results on LINEMOD.} We report GFLOPs and per-object ADD(–S) for various 6D object pose estimation methods. Models found by \textbf{FPG-NAS} achieve superior or competitive performance with lower computational cost. Objects with $^{*}$ are symmetric.}
    \label{tab:linemod_main}
    \begin{adjustbox}{max width=0.9\textwidth}
      \begin{tabular}{c ccccc|ccc}
        \toprule
           & YOLO6D~\cite{Tekin_2018_CVPR} & PVNet~\cite{Peng_2019_CVPR} & HRPose~\cite{guan2022hrpose} & ADLP~\cite{Guo_2023_CVPR} & \textbf{Ours$_{\text{A}}$} 
           & ADLP~\cite{Guo_2023_CVPR} & Lite-HRPE~\cite{10591899} & \textbf{Ours$_{\text{B}}$} \\
        \cmidrule(lr){2-6} \cmidrule(lr){7-9}
        NAS-based?   & \xmark & \xmark & \xmark & \xmark & \textbf{\cmark} & \xmark & \xmark & \textbf{\cmark} \\
        \cmidrule(lr){2-6} \cmidrule(lr){7-9}
        FLOPs [G]    & $$26.1$$ & $$72.7$$ & $$15.5$$ & $$17.3$$ & \textbf{$$10.03$$} 
                    & $$4.8$$  & $$8.9$$  & \textbf{$$4.37$$} \\
        \cmidrule(lr){1-6} \cmidrule(lr){7-9}
        Ape         & $$21.62$$ & $$43.62$$ & $$65.36$$ & $$76.20$$ & \textbf{$$79.92$$} 
                    & \textbf{$$69.90$$} & $$61.21$$ & $$67.56$$ \\
        Benchvise   & $$81.80$$ & \textbf{$$99.90$$} & $$97.38$$ & $$96.70$$ & $$95.45$$ 
                    & $$93.70$$ & \textbf{$$95.53$$} & $$94.10$$ \\
        Cam         & $$36.57$$ & $$86.86$$ & $$85.98$$ & $$92.00$$ & \textbf{$$92.30$$} 
                    & $$84.50$$ & $$84.89$$ & \textbf{$$87.64$$} \\
        Can         & $$68.80$$ & \textbf{$$95.47$$} & $$94.88$$ & $$94.00$$ & $$95.04$$ 
                    & $$83.90$$ & \textbf{$$93.60$$} & $$83.29$$ \\
        Cat         & $$41.82$$ & $$79.34$$ & $$87.33$$ & \textbf{$$88.60$$} & $$87.03$$ 
                    & $$81.60$$ & \textbf{$$86.03$$} & $$83.53$$ \\
        Driller     & $$63.51$$ & $$96.43$$ & \textbf{$$96.73$$} & $$94.80$$ & $$96.63$$ 
                    & $$90.30$$ & \textbf{$$96.23$$} & $$89.01$$ \\
        Duck        & $$27.23$$ & $$52.58$$ & $$71.52$$ & \textbf{$$74.70$$} & $$72.68$$ 
                    & $$68.90$$ & $$68.05$$ & \textbf{$$69.24$$} \\
        Eggbox$^{*}$& $$69.58$$ & \textbf{$$99.15$$} & $$99.06$$ & $$99.30$$ & $$98.78$$ 
                    & $$96.40$$ & \textbf{$$98.97$$} & $$97.19$$ \\
        Glue$^{*}$  & $$80.02$$ & $$95.66$$ & $$97.49$$ & $$97.70$$ & \textbf{$$98.07$$} 
                    & $$93.20$$ & \textbf{$$97.00$$} & $$94.22$$ \\
        Hole-punch  & $$42.63$$ & $$81.92$$ & $$80.55$$ & \textbf{$$82.20$$} & $$81.31$$ 
                    & $$76.30$$ & \textbf{$$78.10$$} & $$75.27$$ \\
        Iron        & $$74.97$$ & \textbf{$$98.88$$} & $$95.90$$ & $$93.20$$ & $$94.17$$ 
                    & $$90.50$$ & \textbf{$$95.50$$} & $$92.54$$ \\
        Lamp        & $$71.11$$ & \textbf{$$99.33$$} & $$97.70$$ & $$96.80$$ & $$95.78$$ 
                    & $$94.60$$ & \textbf{$$96.64$$} & $$93.68$$ \\
        Phone       & $$47.74$$ & \textbf{$$92.41$$} & $$89.91$$ & $$89.60$$ & $$91.86$$ 
                    & $$79.20$$ & \textbf{$$86.65$$} & $$80.14$$ \\
        \cmidrule(lr){1-6} \cmidrule(lr){7-9}
        AVG.        & $$55.95$$ & $$86.27$$ & $$89.21$$ & $$90.40$$ & \textbf{$$90.69$$} 
                    & $$84.85$$ & \textbf{$$87.57$$} & $$85.19$$ \\
        \bottomrule
      \end{tabular}
    \end{adjustbox}
  \end{table*}
\endgroup



\subsection{Results on LINEMOD}
We report in Table~\ref{tab:linemod_main} the performance of the two architectures discovered by \textbf{FPG-NAS} on the LINEMOD dataset, comparing them against several state-of-the-art methods. 
Our two variants, \textbf{Ours$_{\text{A}}$} and \textbf{Ours$_{\text{B}}$}, demonstrate strong ADD(–S) performance, despite their low computational cost. \textbf{Ours$_{\text{A}}$} achieves the highest average accuracy ($90.69$\%) among all baselines in the first group, outperforming methods such as ADLP~\cite{Guo_2023_CVPR}, HRPose~\cite{guan2022hrpose} and PVNet~\cite{Peng_2019_CVPR}, while also being the most efficient with only $10.03$ GFLOPs. \textbf{Ours$_{\text{B}}$} achieves an average accuracy of $85.19$\%, closely following Lite-HRPE~\cite{10591899} ($87.57$\%), but with significantly lower FLOPs ($4.37$G vs. $8.9$G). Notably, \textbf{Ours$_{\text{A}}$} achieves the best results on several challenging objects such as \textit{Ape}, \textit{Cam}, and \textit{Glue}, highlighting the robustness of our models even when compared to the ones with up to seven times the computational cost.

\subsection{Results on SPEED+}
Comparisons between our approach, \textbf{Ours$_\text{A}$}, and baseline methods reporting model FLOPs on the SPEED+ challenge dataset~\cite{park2022speed+} are presented in Table~\ref{tab:speedplus_main}. Notably, our model represents the lowest FLOPs ($10.0$G) while also delivering the best overall accuracy. On the \texttt{Synthetic} and \texttt{Lightbox} subsets, it outperforms others across all metrics. While SPNv2 slightly outperforms under the challenging \texttt{Sunlamp} setting, our method remains competitive. \textbf{Ours$_\text{B}$} was not included here, as no existing baselines at this scale are available for fair comparison.

\begin{table}[H]
  \centering
  \scriptsize
  \setlength{\tabcolsep}{3pt}
  \renewcommand{\arraystretch}{0.7}
  \caption{\textbf{Results on SPEED+.} Pose estimation performance on SPEED+ (\textbf{Ours$_\text{A}$} vs.\ baselines).}
  \label{tab:speedplus_main}
  \resizebox{\columnwidth}{!}{%
    \begin{tabular}{@{}c l c c c@{}}
      \toprule
       & 
      & YOLOv8s-pose~\cite{BECHINI2025198}
      & $^\dagger$SPNv2 ($\phi=0$)~\cite{park2024robust}
      & \textbf{Ours$_\text{A}$} \\
      \cmidrule(lr){3-5}
      & NAS-based?      & \xmark     & \xmark     & \textbf{\cmark} \\
      \cmidrule(lr){3-5}
      & FLOPs [G]       & 29.5       & 12.1       & \textbf{10.0} \\
      \midrule
      \multirow{3}{*}{\texttt{Synthetic}}
        & $E_T$ [m]      & --         & 0.050      & \textbf{0.035} \\
        & $E_R$ [$^\circ$] & --       & 1.441      & \textbf{1.355} \\
        & $E_{\text{pose}}$ & --      & 0.033      & \textbf{0.029} \\
      \midrule
      \multirow{3}{*}{\texttt{Lightbox}}
        & $E_T$ [m]      & 0.758      & 0.447      & \textbf{0.321} \\
        & $E_R$ [$^\circ$] & 18.00    & 16.804     & \textbf{15.303} \\
        & $E_{\text{pose}}$ & 0.432   & 0.368      & \textbf{0.320} \\
      \midrule
      \multirow{3}{*}{\texttt{Sunlamp}}
        & $E_T$ [m]      & 0.518      & \textbf{0.372} & 0.416 \\
        & $E_R$ [$^\circ$] & 19.80    & \textbf{19.366} & 22.823 \\
        & $E_{\text{pose}}$ & 0.432   & \textbf{0.401} & 0.467 \\
      \bottomrule
    \end{tabular}%
  }
  \raggedright
  \scriptsize $^\dagger$Replicated results training heatmap head only.
\end{table}

\subsection{Comparisons with Alternative DNAS techniques}
In Table~\ref{tab:nas_darts_dnal}, our method is compared against DARTS~\cite{liu2018darts} and DNAL~\cite{GUO2022108448}, which adopt \texttt{softmax} and scaled \texttt{sigmoid} weighting strategies, respectively. Under the same FLOPs constraints, our approach consistently outperforms both baselines. This improvement is attributed to our independent gating mechanism, which avoids the early convergence of \texttt{softmax} and the slow activation dynamics of scaled \texttt{sigmoid}, enabling more effective and diverse architecture exploration.


\begin{table}[H]
  \centering
  \scriptsize
  \setlength{\tabcolsep}{2.5pt}
  \setlength{\aboverulesep}{0pt}
  \setlength{\belowrulesep}{0pt}
  \setlength{\extrarowheight}{3pt} 
  \caption{LINEMOD and SPEED+ results at 10\,GFLOPs. Pose error for SPEED+ (lower is better) and average ADD(–S) across all classes for LINEMOD (higher is better).}
  \label{tab:nas_darts_dnal}
  \sisetup{
    table-number-alignment = center,
    table-figures-integer  = 2,
    table-figures-decimal  = 3
  }
  \begin{tabular}{
    l
    S[table-format=2.2]
    S[table-format=2.2]
    S[table-format=2.2]
    S[table-format=1.3]
    S[table-format=1.3]
    S[table-format=1.3]
  }
    \toprule
    \multicolumn{1}{c}{\multirow{2}{*}{Method}} & \multicolumn{2}{c}{\textbf{LINEMOD}~\cite{10.1007/978-3-642-37331-2_42}} & \multicolumn{4}{c}{\textbf{SPEED+}~\cite{park2022speed+}} \\
    \cmidrule(lr){2-3} \cmidrule(lr){4-7}    
      & {FLOPs\,[$\mathrm{G}$]}
      & {ADD(–S)$\uparrow$}
      & {FLOPs\,[$\mathrm{B}$]}
      & {\texttt{Syn.}$\downarrow$}
      & {\texttt{Light.}$\downarrow$}
      & {\texttt{Sun.}$\downarrow$} \\
    \midrule
    DARTS~\cite{liu2018darts}
      & $$10.20$$ & $$85.94$$
      & $$10.08$$ & $$0.046$$ & $$0.356$$ & $$0.529$$ \\
    DNAL~\cite{GUO2022108448}
      & $$10.35$$ & $$87.54$$
      & $$10.16$$ & $$0.036$$ & $$0.339$$ & $$0.509$$ \\
    \textbf{Ours$_{\text{A}}$}
      & $\mathbf{10.03}$ & $\mathbf{90.69}$
      & $\mathbf{10.00}$ & $\mathbf{0.029}$ & $\mathbf{0.320}$ & $\mathbf{0.467}$ \\
    \bottomrule
  \end{tabular}
\end{table}

\section{Conclusion and Future Directions}
\label{sec:conclusion}

In this work, we introduced FPG-NAS, a FLOPs-aware Gated Differentiable Neural Architecture Search framework for efficient 6DoF object pose estimation. Combining a task-specific search space with a differentiable gating mechanism, FPG-NAS discovers architectures that achieve superior accuracy-efficiency trade-offs on LINEMOD~\cite{10.1007/978-3-642-37331-2_42} and SPEED+~\cite{park2022speed+}, outperforming existing methods under strict FLOPs budgets. We observed that multi-candidate selection sometimes activates blocks with overlapping characteristics. Exploring block merging strategies could further streamline the final architectures. Applying FPG-NAS beyond heatmap-based pipelines to tasks such as direct pose regression or dense correspondence-based approaches opens exciting opportunities to broaden its impact across 6DoF pose estimation paradigms.




\section*{Acknowledgment}
This work is supported by the National Research Fund (FNR), Luxembourg, under the C21/IS/15965298/ELITE project and by Infinite Orbits. Experiments were conducted on Luxembourg’s supercomputer MeluXina at LuxProvide.

\printbibliography[heading=bibintoc]

\end{document}